\documentclass[conference, a4paper]{IEEEtran}
\IEEEoverridecommandlockouts
\usepackage{cite}
\usepackage{amsmath,amssymb,amsfonts}
\usepackage{graphicx}
\usepackage{textcomp}
\usepackage{xcolor}
\usepackage{csquotes}
\usepackage{mathtools}
\usepackage{svg}
\usepackage{algorithm}
\usepackage{algpseudocode}
\usepackage{booktabs}
\usepackage{subfig}
\usepackage{microtype}
\def\BibTeX{{\rm B\kern-.05em{\sc i\kern-.025em b}\kern-.08em
    T\kern-.1667em\lower.7ex\hbox{E}\kern-.125emX}}
\makeatletter \newcommand{\linebreakand}{\end{@IEEEauthorhalign}
  \hfill\mbox{}\par
  \mbox{}\hfill\begin{@IEEEauthorhalign}
}
\makeatother \DeclareRobustCommand{\IEEEauthorrefmark}[1]{\smash{\textsuperscript{\footnotesize #1}}}
\renewcommand{\Re}[1]{\ensuremath{\operatorname{Re}(#1)}}
\renewcommand{\Im}[1]{\ensuremath{\operatorname{Im}(#1)}}

\definecolor{rwth-blue}{cmyk}{1,.5,0,0}\colorlet{rwth-lblue}{rwth-blue!50}\colorlet{rwth-llblue}{rwth-blue!25}
\definecolor{rwth-violet}{cmyk}{.6,.6,0,0}\colorlet{rwth-lviolet}{rwth-violet!50}\colorlet{rwth-llviolet}{rwth-violet!25}
\definecolor{rwth-purple}{cmyk}{.7,1,.35,.15}\colorlet{rwth-lpurple}{rwth-purple!50}\colorlet{rwth-llpurple}{rwth-purple!25}
\definecolor{rwth-carmine}{cmyk}{.25,1,.7,.2}\colorlet{rwth-lcarmine}{rwth-carmine!50}\colorlet{rwth-llcarmine}{rwth-carmine!25}
\definecolor{rwth-red}{cmyk}{.15,1,1,0}\colorlet{rwth-lred}{rwth-red!50}\colorlet{rwth-llred}{rwth-red!25}
\definecolor{rwth-magenta}{cmyk}{0,1,.25,0}\colorlet{rwth-lmagenta}{rwth-magenta!50}\colorlet{rwth-llmagenta}{rwth-magenta!25}
\definecolor{rwth-orange}{cmyk}{0,.4,1,0}\colorlet{rwth-lorange}{rwth-orange!50}\colorlet{rwth-llorange}{rwth-orange!25}
\definecolor{rwth-yellow}{cmyk}{0,0,1,0}\colorlet{rwth-lyellow}{rwth-yellow!50}\colorlet{rwth-llyellow}{rwth-yellow!25}
\definecolor{rwth-grass}{cmyk}{.35,0,1,0}\colorlet{rwth-lgrass}{rwth-grass!50}\colorlet{rwth-llgrass}{rwth-grass!25}
\definecolor{rwth-green}{cmyk}{.7,0,1,0}\colorlet{rwth-lgreen}{rwth-green!50}\colorlet{rwth-llgreen}{rwth-green!25}
\definecolor{rwth-cyan}{cmyk}{1,0,.4,0}\colorlet{rwth-lcyan}{rwth-cyan!50}\colorlet{rwth-llcyan}{rwth-cyan!25}
\definecolor{rwth-teal}{cmyk}{1,.3,.5,.3}\colorlet{rwth-lteal}{rwth-teal!50}\colorlet{rwth-llteal}{rwth-teal!25}
\definecolor{rwth-gold}{cmyk}{.35,.46,.7,.35}
\definecolor{rwth-silver}{cmyk}{.39,.31,.32,.14}
 \begin{document}
\bibliographystyle{IEEEtran}

\title{End-to-End Reinforcement Learning of Curative Curtailment with Partial Measurement Availability
\thanks{This work was funded by E.ON SE in cooperation with E.ON Group Innovation GmbH and Schleswig-Holstein Netz AG, who provided the grid topology.
}
}

\author{
\IEEEauthorblockN{
Hinrikus Wolf$^{*\dagger}$\IEEEauthorrefmark{1}, 
Luis Böttcher$^*$\IEEEauthorrefmark{2}, 
Sarra Bouchkati$^*$\IEEEauthorrefmark{2}, 
Philipp Lutat\IEEEauthorrefmark{2},\\
Jens Breitung\IEEEauthorrefmark{1},
Bastian Jung\IEEEauthorrefmark{3},
Tina Möllemann\IEEEauthorrefmark{3}, 
Viktor Todosijević\IEEEauthorrefmark{3}, \\
Jan Schiefelbein-Lach\IEEEauthorrefmark{4},
Oliver Pohl\IEEEauthorrefmark{5},  
Andreas Ulbig\IEEEauthorrefmark{2}, and
Martin Grohe$^\dagger$\IEEEauthorrefmark{1}
}
  \IEEEauthorblockA{\IEEEauthorrefmark{1} Computer Science, RWTH Aachen University \\
  \{hinrikus, grohe\}@informatik.rwth-aachen.de}
  \IEEEauthorblockA{\IEEEauthorrefmark{2} IAEW, RWTH Aachen University \\
   \{{l.boettcher, s.bouchkati, a.ulbig}\}@iaew.rwth-aachen.de
   }
  \IEEEauthorblockA{\IEEEauthorrefmark{3} RWTH Aachen University} 
  \IEEEauthorblockA{\IEEEauthorrefmark{4} E.ON Group Innovation GmbH}
  \IEEEauthorblockA{\IEEEauthorrefmark{5} Schleswig-Holstein Netz AG}
}

\IEEEoverridecommandlockouts
\IEEEpubid{\makebox[\columnwidth]{979-8-3503-9042-1/   24/\$31.00~\copyright2024 IEEE \hfill} 
\hspace{\columnsep}\makebox[\columnwidth]{ }}

\maketitle
{
\def\thefootnote{*}\footnotetext{These authors contributed equally to this work}
}
{
\def\thefootnote{\dag}\footnotetext{Funded by the European Union (ERC, SymSim, 101054974). Views and opinions expressed are, however, those of the author(s) only and do not necessarily reflect those of the European Union or the European Research Council. Neither the European Union nor the granting authority can be held responsible for them.}
}

\begin{abstract}
In the course of the energy transition, the expansion of generation and consumption will change, and many of these technologies, such as PV systems, electric cars and heat pumps, will influence the power flow, especially in the distribution grids. Scalable methods that can make decisions for each grid connection are needed to enable congestion-free grid operation in the distribution grids.
This paper presents a novel end-to-end approach to resolving congestion in distribution grids with deep reinforcement learning (RL). Our RL-agent learns to curtail power and set appropriate reactive power to determine a non-congested and, thus, feasible grid state. 
State-of-the-art methods, such as the optimal power flow (OPF), require high computational costs and detailed measurements of every bus in a grid.
In contrast, the presented method enables decisions under sparse information with just some buses observable in the grid. 
Distribution grids are generally not yet fully digitized and observable, so this method can be used for edge decision-making on the majority of low-voltage grids. On a real low-voltage grid, the approach resolves 100\%  of violations in the voltage band and 98.8\% of asset overloads.
The results show that decisions can also be made on real grids that guarantee sufficient quality for congestion-free grid operation. 
\end{abstract}

\begin{IEEEkeywords}
Curative Curtailment, Edge Computing, State Estimation, Optimal Power Flow, Deep Reinforcement Learning
\end{IEEEkeywords}

\section{Introduction}

As part of the decarbonization of energy systems, new loads and generators are being connected to the distribution grids. The majority of electricity generation plants, such as photovoltaic are being built decentrally. Furthermore, heat pumps and electromobility are replacing technologies that were previously based on fossil fuels. This changes and increases the power flow in the electricity grids, especially in the distribution grids. To cope with this issue, the power demand or generation can be curtailed in cases of congestion to maintain the grids physical limits.
Independently of regulatory constraints, the need to decide which appliances connected to the grid must be curtailed to resolve the congestion exists. Solving an Optimal Power Flow (OPF) is the state-of-the-art approach to this question.
However, to solve an OPF, measurements at every bus in the grid are needed. Often, such measurements are not available. A standard way of coping with missing measurements is to use state estimation techniques to predict the grid state on the basis of a few measurements and then apply optimal power flow to the predicted grid state. Current developments show that predicting grid states by using state estimation is a measure for distribution system operators to obtain complete information about their grid states.
Unfortunately, state estimation is computationally expensive and introduces some inaccuracy.

Our approach shows a novel method for obtaining curative curtailment decisions in power grids in an end-to-end fashion. Based on a few measurements, we utilize a model to predict curtailment decisions with high accuracy, which is deployable on low-cost edge computing devices. Our method uses the possibilities of reinforcement learning with partial observability by iteratively learning the curtailment decisions to obtain non-critical grid states.
Reinforcement learning under partial observability finds normally application, e.g. in robotics \cite{zhao2023robot}.

\subsection{Related work}

Recent studies have demonstrated an increasing interest in utilizing Machine Learning (ML) techniques to estimate the solution of AC-OPF. This is achieved either by supervised learning, through self-supervised approaches or using Reinforcement Learning (RL) paradigms.
Some of the early works relied on constrained supervised training. The work \cite{nellikkath2021physicsinformed} incorporated the Karush-Kuhn-Tacker (KKT) conditions as penalty terms during training, while \cite{fioretto2020lagrangian} combines a Multi-Layer Perceptron (MLP) with the Lagrangian Dual method. The effort in \cite{Owerko.2019} leverages graph learning and implements a local and a global Graph Neural Network (GNN) to solve the OPF problem. 
Although supervised approaches may yield promising results, they require the generation of pre-solved instances for training, which may be time-consuming and resource-intensive. Other approaches mitigate this drawback by relying on self-supervised training.
The work DC3 presented in \cite{dc3donti} implements an MLP, where a completion step based on Newton's Method is performed, and feasibility of the inequality constraints is ensured using a gradient-based correction step. 
On the other hand, other works exploit the self-learning ability of RL-agents and employ similar techniques for constrained learning to direct the RL agent in making physics-informed decisions. \cite{yan2020} and \cite{TongRL} rely on adding system-specific Lagrangian parameters to the reward function, whereas the works in \cite{sayedRLFeasibility} and in \cite{SayedRL2} use the safety convex layer and holomorphic embedding based layer respectively to ensure feasibility. \\
A common thread among these approaches is the assumption of a fully observable grid state. However, in real-world scenarios, grids often lack full observability, with not all measurements from all grid participants readily accessible. This aspect presents an opportunity for developing a new method which takes reduced grid observability into consideration.

\subsection{Contribution}
This paper introduces a novel approach that leverages deep reinforcement learning to determine the level of curtailment for controllable buses, relying on a partially observable grid state. The method utilizes available grid measurements as inputs to a reinforcement learning agent, which directly determines the curtailment level for each bus. The agent undergoes training, following the typical reinforcement learning paradigm, within an environment that simulates the grid under the current supply task using power flow computations. To validate the proposed method, real low-voltage grid data from Schleswig-Holstein Netz AG and synthetic data representing realistic supply tasks are employed for training and validation. Results showcase the model's efficacy in resolving nearly all instances of violations of physical constraints. Instances where violations persist are correctly identified by the model, however, with insufficient curtailment.

 \section{Method}
\label{sec:method}

This section outlines the methodologies used to develop and train a model for grid curtailment with partial measurement availability. The model is capable of handling various operational scenarios ranging from non-critical to critical grid states. Central to our approach is the generation of a realistic dataset presented in \ref{subsec:data_generation} consisting of a wide spectrum of grid states and the application of a reinforcement learning approach presented in \ref{subsec:RLarchitecure} to determine the curtailed power in critical situations.

The primary objective is to maintain the grid within permissible grid states by dynamically adjusting power outputs at controllable nodes within the grid. This involves both the direct control of power generation and the curtailment of power usage to prevent or resolve potential violations such as asset overloads and voltage band deviations. To achieve this, our method relies on a detailed environment that models the grid's physical and operational constraints and employs an actor-critic architecture to optimize the decision-making process in an end-to-end concept.

\subsection{Data Generation}
\label{subsec:data_generation}

To create datasets, it is necessary to map different grid operating states on which decisions have to be made. 
The dataset should consist of both non-critical and critical operational states to enable effective learning during training. By including a variety of scenarios, ranging from normal to critical states, the model can determine patterns and distinctions between the two. This ensures that the detection of critical states can be informed by the characteristics of non-critical states, facilitating a comprehensive learning of grid dynamics.
The second aspect is the mapping of curtailment decisions in the grid so that flexible assets in the grid are curtailed in such a way that a non-critical grid state can be created from a critical grid state by the model's decision. 

\subsubsection{Grid State Modelling}
Since, in many cases, grid models do not contain detailed information on the supply task, the first step is to assign an annual time series to the grid model. These are based on the technologies installed at the nodes, such as photovoltaics and household loads \cite{vertgewall2022modeling, meinecke2020simbench}.
The \textit{controllable} installed technologies are assigned realistic cost coefficients $c$ (cf. Eq. \ref{eq:opf_obj}) so that their optimal use is incited. 
In addition, \textit{observable} and \textit{non-observable} nodes are defined in the grid, which should represent nodes in reality with measurement technology. Based on the modelled supply task on the grid, an optimal power flow (cf. \ref{OPF}) on the grid is calculated. 

\subsubsection{Optimal Power Flow}
\label{OPF}
The OPF used in this method combines the economic dispatch of generation and load and the physical constraints through the power flow formulations. 
Optimal power flow is a nonlinear and non-convex, \textsf{NP}-complete, constrained optimization problem on power grids, commonly used in areas such as grid planning, power markets, and active grid operation and is a backbone in many calculations used to operate modern power systems \cite{cain2012history,bienstock2019strong}. For that reason, there exist different types and modifications of the standard AC OPF problem, e.g., DC OPF, security-constrained OPF, and others \cite{frank2016introduction}. 
The following formulation describes the Standard OPF problem.
Let \(\mathcal{G} = (\mathcal{V}, \mathcal{E})\) be a grid with buses \(\mathcal{V}\) and lines \(\mathcal{E}\).
\begin{subequations}
\begin{alignat}{2}
& \min_{y}        &\quad&  \sum_{i=0}^{|\mathcal{V}|-1} \sum_{k=0}^{n_c} c_{ik} (P_{g_i}^k) \label{eq:opf_obj} \\
& \text{subject to} &      & S_{\text{bus}}(V_m, \Theta) - S(P_g, Q_g) = 0, \label{eq:opf_equal}\\
&&& P_{g}^{\min} \leq P_g \leq P_{g}^{\max}, \label{eq:opf_node_p} \\
&&& Q_{g}^{\min} \leq Q_g \leq Q_{g}^{\max}, \label{eq:opf_node_q} \\
&&& V_{m}^{\min} \leq V_m \leq V_{m}^{\max}, \label{eq:opf_node_v} \\
&&& |S_f(V_m, \Theta)| \leq S^{\max}, \label{eq:opf_edge_sf} \\
&&& |S_t(V_m, \Theta)| \leq S^{\max}, \label{eq:opf_edge_st}
\end{alignat}
\end{subequations}
with \( \Theta = \arg(V) \), \( V_m = |V| \), \( P_g = \Re{S_g} \), and \mbox{\( Q_g = \Im{S_g} \)}. Additionally, \( c \in \mathbb{R}^{|\mathcal{V}|\times n_c} \), \( n_c \in \mathbb{N} \) is the polynomial cost coefficient matrix; \( S_{\text{bus}} \in \mathbb{C}^{|\mathcal{V}|} \) and \( S_f, S_t \in \mathbb{C}^{|\mathcal{E}|} \) represent the complex bus and branch power injections. \( P_{g}^{\min} \), \( P_{g}^{\max} \), \( Q_{g}^{\min} \), \( Q_{g}^{\max} \), \( V_{m}^{\min} \), \( V_{m}^{\max} \) \(\in \mathbb{R}^{|\mathcal{V}|}\) and \( S^{\max} \in \mathbb{R}^{|\mathcal{E}|} \) represent the different upper or lower constraints of the variables. OPF consists of three parts: equation \ref{eq:opf_obj} is the objective function of the minimization problem and represents the monetary cost of operating a power grid in the given state; equation \ref{eq:opf_equal} is the equality constraint of the problem, and represents the physical law of energy conservation; finally, equations (\ref{eq:opf_node_p}--\ref{eq:opf_node_v}) are the node-, and (\ref{eq:opf_edge_sf}--\ref{eq:opf_edge_st}) edge-level inequality constraints, representing different technical limits of grid operation.

\subsubsection{Training Data Set}

The calculation is carried out for at least 35.040 quarter-hour values in order to represent an annual simulation \cite{pandapower.2018, matpower}. In the course of development, it has been shown that further augmentation (cf. \ref{augmentaion}) of the reference time series on the basis of the annual simulation is reasonable, in particular, to increase the number of critical grid states in the dataset in order to achieve a higher accuracy of the trained model.
Furthermore, when calculating the optimal power flow, the grid is assumed to be completely \textit{observable}. For comparison, a state estimation is also calculated in combination with an optimal power flow in order to take account of the incomplete observability and to obtain a reference for the model to be trained here. 

Therefore, the training data set consists of tuples of unsolved and solved optimal power flow as well as an extension by state estimation and optimal power flow on reduced observability. The data set generation, therefore, follows a similar approach as presented in \cite{GNN_PF_2023}.

\subsubsection{Augmentation}
\label{augmentaion}
To enlarge our training set, we performed data augmentation, where we added noise to the bounds for \(P^\text{min}\), \(P^\text{max}\).
The set points of each augmented bus are set to its maximum.
We augment cases with violations of the lower voltage band to increase their representation to ensure a more balanced distribution of violations in the dataset. We keep the non-augmented initial dataset as test data and train only on the augmented data.

\subsection{Reinforcement Learning Methodology}
\label{subsec:RLarchitecure}
\begin{figure}[h]
    \centering
    \includegraphics[width=\columnwidth]{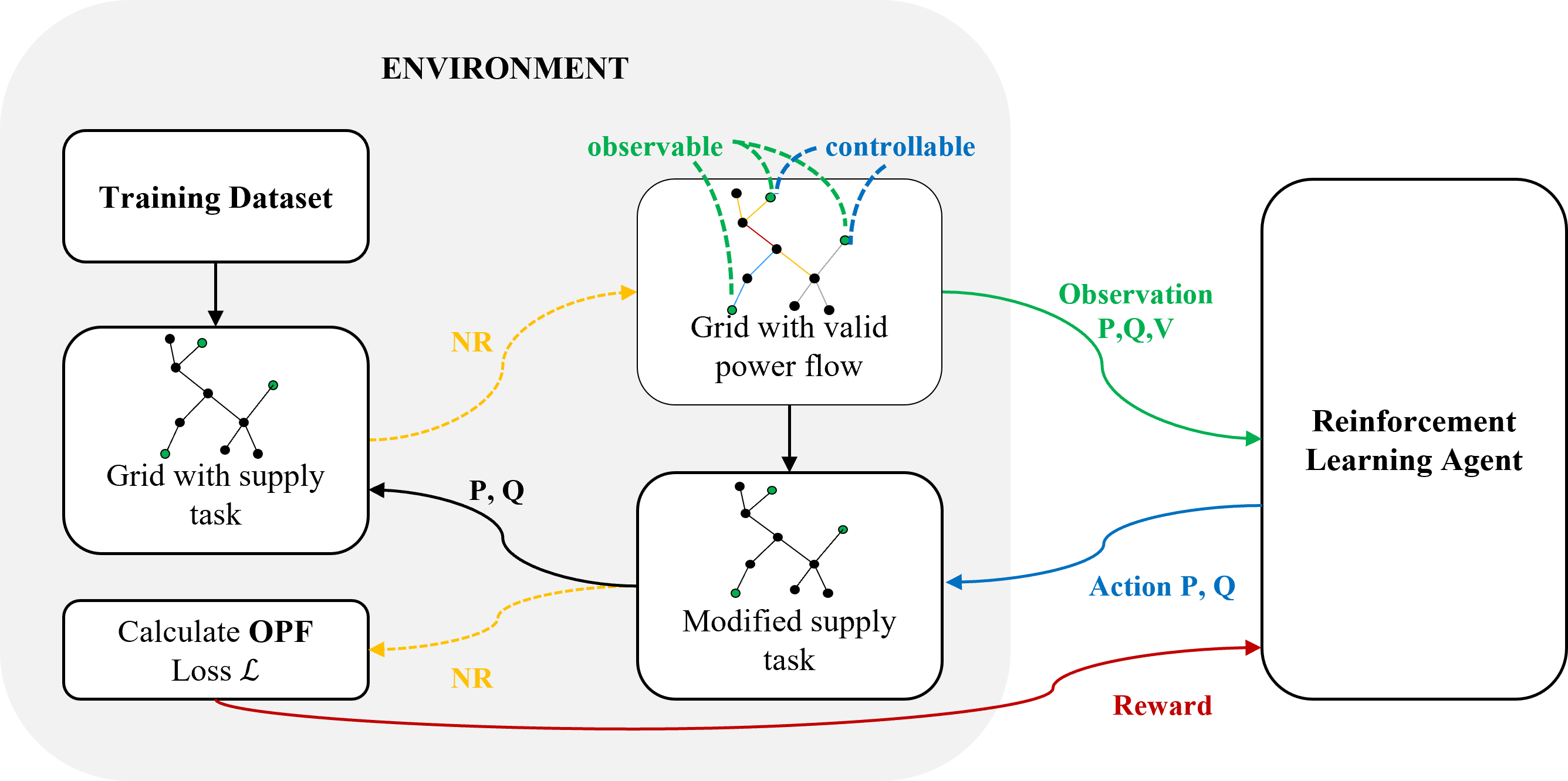}
    \caption{Overview of Model Methodology}
    \label{fig:architecture}
\end{figure}

A reinforcement learning approach has been developed for end-to-end learning of grid operation decisions, which takes into account the incomplete availability of measurement data in electricity grids. We call this approach grid decision learning (GDL). 
The methodology for training the reinforcement learning (RL) agent is shown in Fig. \ref{fig:architecture}.

The environment has access to all physical properties of the grid. It is predefined which nodes have controllable assets and which do not. 
The logic works as presented in the following.

\begin{algorithm}
\label{alg:RL}
\caption{Reinforcement Learning Training Procedure}
\begin{algorithmic}[1]
    \State Define loads and generation for each node from the data set for each time step.
    \State Use the Newton-Raphson solver to obtain the current grid state, simulating real-world physics.
    \State Generate an observation from the valid grid state, including $P$, $Q$, and $V$ at controllable nodes.
    \State RL agent receives the observation and determines the action vector.
    \State Apply the action to the environment; recompute grid state using Newton-Raphson.
    \State Calculate the reward based on voltage band violations, line overloading, and curtailment costs.
    \State Move to the next time step, maintaining the curtailment setting.
    \State Pass the reward and the observation back to the agent.
\end{algorithmic}
\end{algorithm}

The architecture is designed to adequately learn the curative curtailment decisions on a single grid topology. This is done by resolving grid violations by curtailing active \(P\) and reactive power \(Q\) at certain buses with flexibility, which are explicitly listed as \emph{controllable}.
Furthermore, the agent is only provided with measurements \(P, Q\) and \(V\) of \emph{observable} buses. 
We assume that every controllable bus is also observable. Additionally, we supply valid ranges \([P_i^\text{min}, P_i^\text{max}]\) and \([Q_i^\text{min}, Q_i^\text{max}]\) for each controllable bus \(i\).
The reinforcement learning agent shall learn the optimal setpoint for \(P\) and \(Q\) at the controllable buses based on the restricted measurements (or observation).
The agent that determines the \(P\) and \(Q\) setpoints is a neural network obtained from an Actor-Critic model, which we train using Deep Deterministic Policy Gradient (DDPG) \cite{lillicrap2016continues}.
Both Actor and Critic are simple multi-layer perceptrons with two hidden layers, each of width 512. 
We have implemented the model in PyTorch \cite{PyTorch} with TorchRL \cite{TorchRL}.

\subsubsection{Environment}
The environment can be interpreted as a simulation of the physical world.
It is provided with a fixed grid topology from the data set generation process, which gets updated in every training step.
The agent interacts with the environment by collecting observations at all observable buses. Based on this observation, the agent learns to apply an action by determining relative $P$ and $Q$ setpoints, which are provided to the environment and update the setpoints in the grid topology. A power flow is computed with a Newton-Raphson procedure to simulate the effect of the agent's decision. Based on the overall resulting state of the environment, which includes measurements of buses not observable by the agent, a reward is given to the agent.

\subsubsection{Reward}
The reward evaluates the agent's action on the basis of complete information about the grid status. 
Let $n$ be the number of buses, $k$ the number of controllable generators and $\ell$ be the number of lines.
Let \(\mathbf{V}\) be the vector of voltages of the buses in the current state, and denote by \(\mathbf{V}^\text{min}\) and \(\mathbf{V}^\text{max}\) the vector of lower and upper limits for the voltage band. We define the voltage loss $\mathcal{L}_V \coloneqq \max_{i \in \{1, \ldots, n\}} |\mathbf{V} - \text{clamp}(\mathbf{V}, \mathbf{V}^\text{min}, \mathbf{V}^\text{max})|_i$ as the maximum deviation from the voltage band. 
Formally, for vectors \(\mathbf{x}, \mathbf{a}, \mathbf{b} \in \mathbb{R}^{n}\) we define $\text{clamp}(\mathbf{x}, \mathbf{a}, \mathbf{b}) \coloneqq \max(\mathbf{a}, \min(\mathbf{x}, \mathbf{b}))$ where \(\min\) and \(\max\) are applied pointwise.
Similarly, for the vector of relative loads $\mathbf{I}$ we let $\mathcal{L}_I \coloneqq \max_{i \in \{1, \ldots, \ell\}} \max(\mathbf{I} - \mathbf{1}, \mathbf{0})_i$ be the maximum relative line load violation.
Further, we define the curtailment cost as $\mathcal{C}_P \coloneqq \frac{1}{|\mathbf{C}|} \sum_{i=1}^{|\mathbf{C}|} \mathbf{C}_i$ where $\mathbf{C}_i$ is the amount of power curtailed at the $i$-th controllable generator.
The agent obtains a negative reward if the power flow (computed after applying the agent's actions) does not converge. Otherwise, the agent obtains
\[
    \mathcal{R} \coloneqq \begin{cases}
-\min\left(\frac{\mathcal{L}_{V}}{s} + \mathcal{L}_I, 1\right) & \text{if $\mathcal{L}_V + \mathcal{L}_I> 0$}, \\
        1-\frac{\mathcal{C}_{P}}{s} & \text{otherwise,} 
    \end{cases}
\]
where $s \coloneqq \frac{\lambda}{k} \cdot \sum_{i=1}^{k}|\mathbf{P}^\text{max}_i - \mathbf{P}^\text{min}_i|$ for a $\lambda > 1$.
Since $\mathcal{C}_P \leq s$, we have $\mathcal{R} \in [-1, 1]$ if the Power Flow converges. Importantly, any action that results in no violations is strictly better than all actions that lead to violations. 
This is done to incentivize the agent to resolve critical grid states.
However, this reward is not continuous due to the step of size $s$ when moving from states with violations to those without.
Note that the reward comprises information not available to the agent, specifically voltage and line load violations at buses that are not observable.

\subsection{Training Procedure}

Experiences are collected in a Replay Buffer of size  200,000 from which we sample 100 individual experiences uniformly at random at a time, which are then used to train the Actor and Critic networks.
\\

\section{Experiments}
\label{experiments}
In this section, we present the experimental results of our method on an example grid from reality. First, we analyse the dataset in \ref{subsec:dataset_exp} and describe our training process in \ref{subsec:training_exp} before evaluating the results in \ref{subsec:eval_exp}.
\\

\subsection{Data Set Generation}
\label{subsec:dataset_exp}
\begin{figure}[t]
    \centering
    \includegraphics[width=0.9\columnwidth]{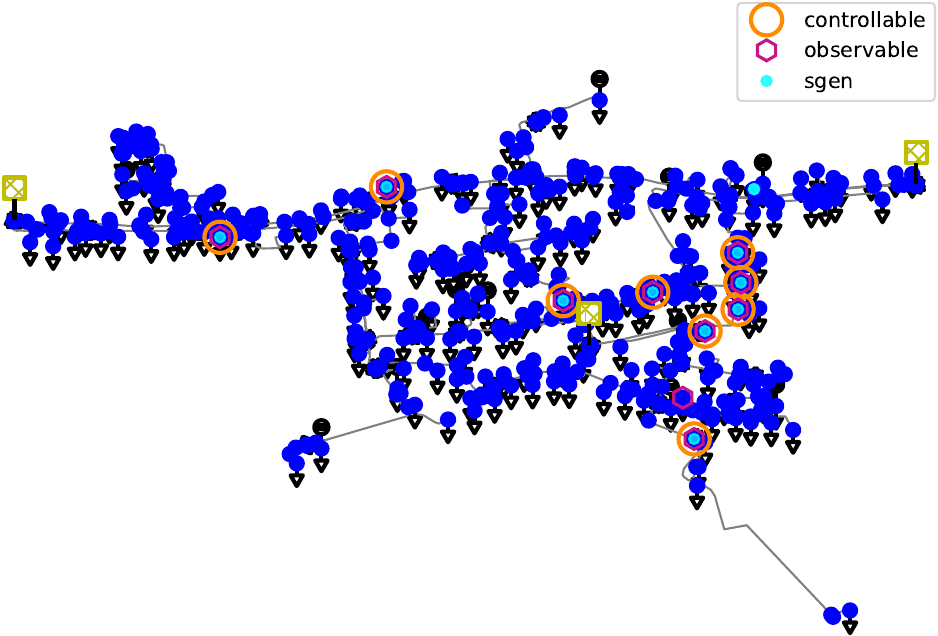}
    \caption{Low Voltage grid with 7\% observability (provided by Schleswig Holstein Netz AG)}
    \label{fig:lv_eon_grid}
\end{figure}

We conduct our experiments on a real low voltage grid of Schleswig-Holstein Netz AG, to which we apply our data generation method described in Section \ref{subsec:data_generation} with synthetic load and generation profiles. We defined 7\% observable and controllable nodes in the grid as depicted in Figure \ref{fig:lv_eon_grid}.\\

Table \ref{tab:data_set_stats} presents the distribution of violations in the generated data set (size:  20,453 quarter-hour supply tasks) and the ratio of resolved violations. Note that violations in either category are non-exclusive.

\begin{table}[h]
    \centering
     \caption{Distribution of violations in experiment dataset}
    \label{tab:data_set_stats}
    \begin{tabular}{lrr}
    \toprule
        Violations & Number   & Solved \\\midrule
        Total & 7,041 & 99.7\% \\
        Upper Voltage Band & 5,384 & 100\% \\
        Lower Voltage Band & 424 & 100\%\\
        Asset Overloaded & 6,617 & 98.8\% \\\bottomrule
    \end{tabular}
   
\end{table}
\subsection{Training}
\label{subsec:training_exp}

We train on shuffled data in batches of each 20 supply tasks.
The model gets each time-step five times in a row.
Except for the first time, the set points were adapted according to the model's previous decision, so the model has the chance to correct previous decisions.
We trained for 1,200,000 steps, so 240,000 different training samples (of augmented data), each five times.
We run our experiments on a desktop computer with an AMD Ryzen 7 7700X processor and an NVIDIA GeForce RTX 4070 Ti GPU.

\subsection{Evaluation}
\label{subsec:eval_exp}
To evaluate the models, we compare the results of the RL agent (\texttt{rl}) with the OPF (\texttt{opf}). 
However, complete information is required for the OPF, so a state estimation is necessary before using it. 

In the following figures, test results of our model in comparison to the \texttt{opf} are depicted.
On the \(x\) axes, the \emph{relative curtailment} of the current grid state is visualised.
The relative curtailment is the ratio of the sum of absolute power output or input, respectively, at each controllable bus and the maximal flexibility.
The colour of each point corresponds to the available flexibility available in the grid in kW. Each point in the plots represents a single supply task on the grid topology and shows the worst-case situation on the grid (highest asset loading or lowest/highest voltage).

\begin{figure}[h]
    \centering
    \includegraphics[width=\columnwidth]{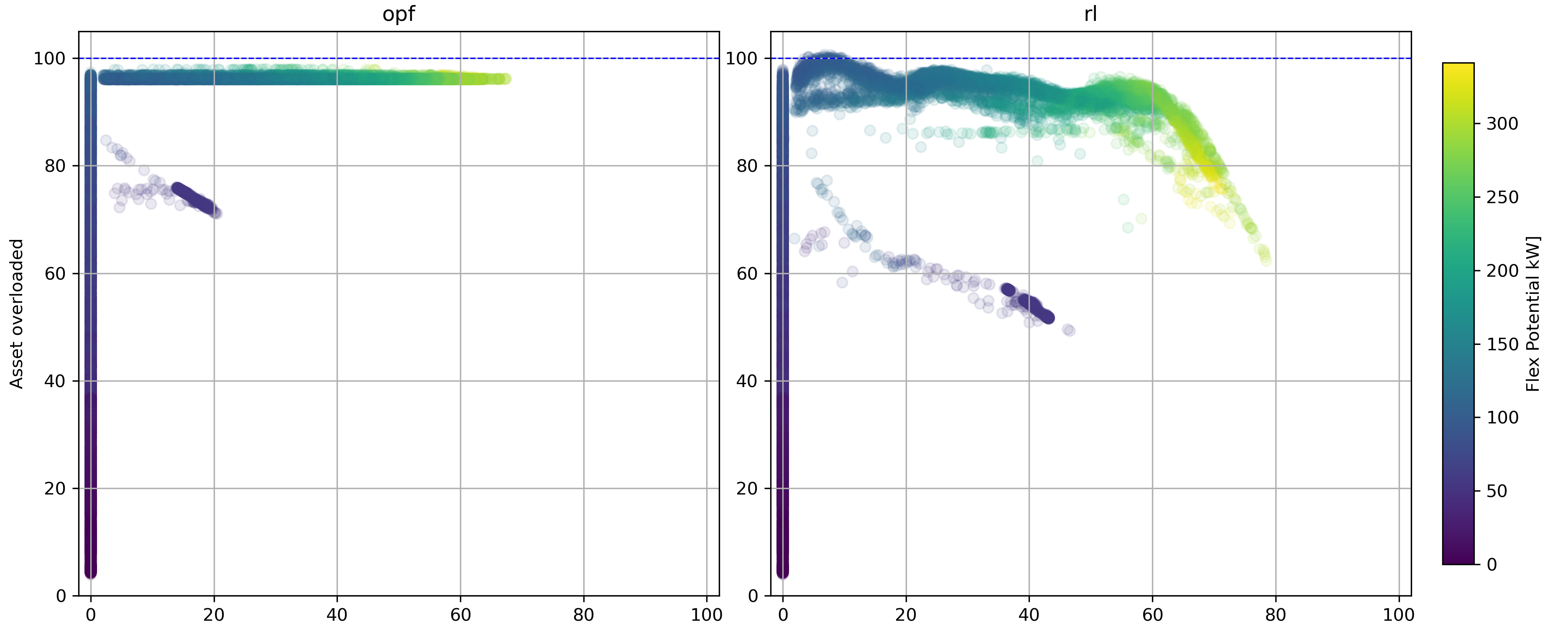}
    \caption{Asset Overloading and Relative P Curtailment}
    \label{fig:P-scatter}
\end{figure}

In Fig. \ref{fig:P-scatter}, the asset overloading in the grid is presented. The \texttt{opf} solution presents an optimal curtailment of all flexibilities to maintain the grid within the operational limits. The curtailment is along the 100\% line to solve all critical asset loadings cost optimal. The \texttt{rl} decision shows similar behaviour but in a suboptimal way. The agent's decisions are, in a way, overachieving and curtailing increasingly more to stay within the permissible physical bounds but with way less available data than the OPF. 

\begin{figure}[h]
    \centering
    \includegraphics[width=\columnwidth]{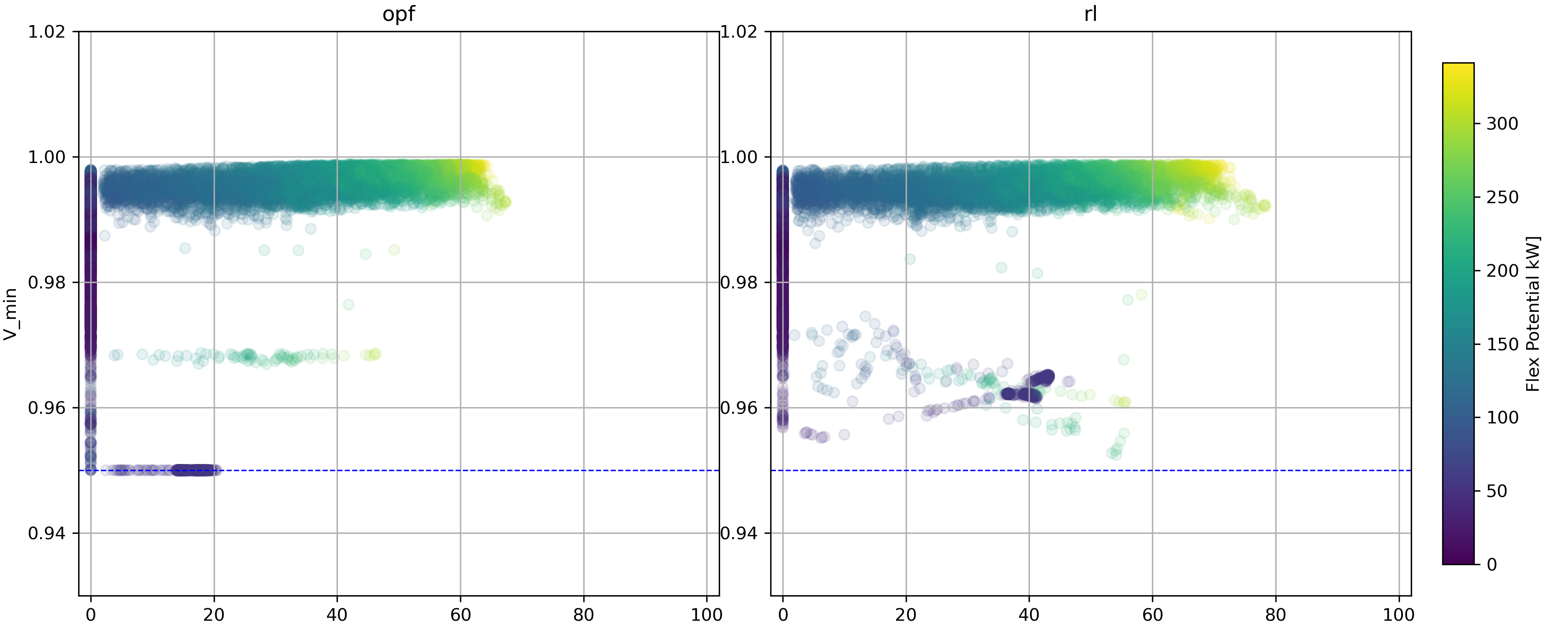}
    \caption{Lower Voltage Band and Relative P Curtailment}
    \label{fig:Vmin-scatter}
\end{figure}

When looking at the lower voltage band in Fig. \ref{fig:Vmin-scatter}, there are few curtailment decisions done by the \texttt{opf} to stay within the voltage band of 0.95 p.u. and 1.05 p.u. 
The \texttt{rl} also stays within the voltage band, but a similar suboptimal decision can be observed. 

\begin{figure}[h]
    \centering
    \includegraphics[width=\columnwidth]{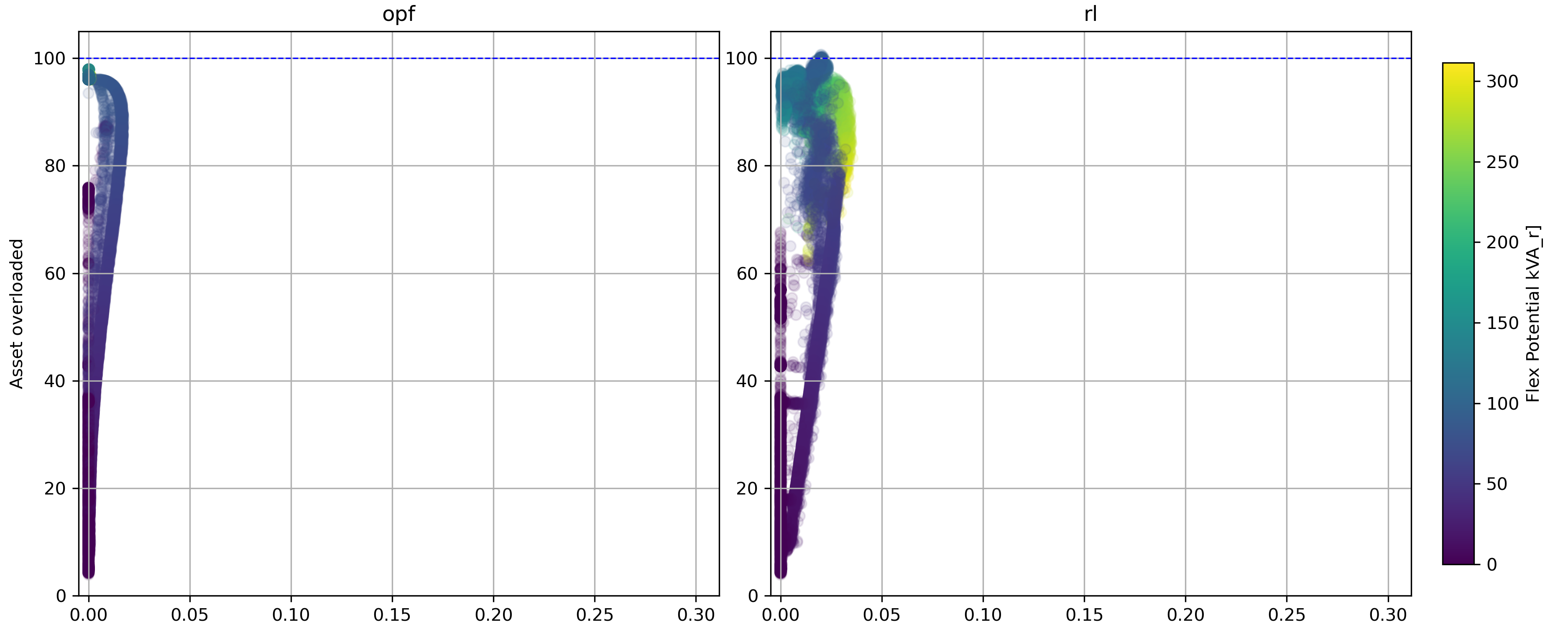}
    \caption{Asset Overloading and Relative Q Curtailment}
    \label{fig:Q-scatter}
\end{figure}

Lastly, we looked at the relative Q curtailment in Fig. \ref{fig:Q-scatter} to compare the decisions of the \texttt{rl} agent with the \texttt{opf}. It is notable that the \texttt{rl} agent learned to use $Q$ similarly as the \texttt{opf} to achieve congestion-free grid states.

Except for 22 cases in the asset overload, all violations have been resolved by the \texttt{rl} agent. Though the \texttt{rl} agent detects these 22 cases as violations, it does not curtail them sufficiently. The \texttt{rl} agent shows much more noise and more curtailment but learns to curtail with just limited information with a high accuracy and high detection rate of critical grid states.
When comparing the computation times \ref{tab:computation-times}, the end-to-end learning approach outperforms the state of the methods.

\begin{table}[h]
    \centering
    \caption{Average Computation and Training Times}
    \begin{tabular}{ccccc}
    \toprule
         Training (total) &  Inference (per supply task) &  OPF & OPF+SE\\
         \midrule
         8h & 0.09s & 0.31s & 51s \\
    \bottomrule
    \end{tabular}
    
    \label{tab:computation-times}
\end{table}

Especially when comparing the \texttt{rl} agent with the OPF with an initial State Estimation, the computation time increases significantly.

 \section{Discussion}

The results show that the model behaves similarly to the OPF, probably owing to the design of the reward function, which penalises violations and encourages minimal curtailment, a feature shared with traditional OPF solvers. Nevertheless, the model still tends to curtail a higher amount of power compared to the OPF. In addition, a correlation between the amount of curtailment and the flexibility available in the system is observable. Test cases in which violations persist after the agent's action indicate that the agent responds appropriately, but falls short of curtailing power sufficiently. This shortfall is observed for lower voltage band violations and may be due to their under representation in the training data. On the other hand, the model occasionally opts for curtailment even in the absence of detected violations, a behaviour observed in 25\% more cases compared to the OPF. However, the magnitude of the curtailment in such cases remains minimal. Notably, in terms of computational efficiency, the model exhibits an inference time almost four times faster than the OPF and approximately 566 times faster than combined state estimation with the OPF. This significant acceleration renders the method highly attractive for real-time applications.

 \section{Conclusion}
We introduce a machine learning model for solving state estimation and OPF in an end-to-end fashion. This grid decision learning enables the curative curtailment of flexibilities.
We train and validate our model on real grid data and show that the model is able to detect and counteract violations in most cases, while being faster than traditional solvers.
Although the model does not provide sufficient curtailment in some cases, it correctly detects all violations.
This drawback could be mitigated with more training data and further fine-tuning of the hyperparameters.

For future work, the model could be extended to quantify the uncertainty of its output.
In grid operation, the models' decisions are only applied if the model is certain of its decision.
The overall accuracy could be improved by using ensembles of embeddings, where the outputs are ranked by their (un)certainty.

\end{document}